\pdfoutput=1
\documentclass[11pt]{article}
\PassOptionsToPackage{dvipsnames}{xcolor}
\usepackage[final]{acl}

\usepackage[utf8]{inputenc}
\usepackage[T1]{fontenc}

\usepackage{multirow}

\usepackage{times}
\usepackage{latexsym}
\usepackage{amsmath}

\usepackage{booktabs} % For professional-looking tables
\usepackage{graphicx} % For image inclusion if needed

\usepackage{microtype}

\usepackage{inconsolata}

\usepackage{graphicx}

\usepackage{adjustbox}
\usepackage{booktabs}

\usepackage{tabularx}  % For adjustable column widths
\usepackage{multirow}  % For multi-row cells
\usepackage{caption}

\title{Disparities in LLM Reasoning Accuracy and Explanations: A Case Study on African American English }
\newcommand{\aspace}{\hspace{2em}}
\newcommand{\cmu}{$^\heartsuit$}
\newcommand{\ucla}{$^\clubsuit$}
\newcommand{\uva}{$^\diamondsuit$}

\author{
Runtao Zhou\uva \thanks{Equal contribution.} \aspace
Guangya Wan\uva \footnotemark[1] \aspace Saadia Gabriel\ucla
\\ \textbf{Sheng Li\uva \aspace Alexander J Gates\uva \aspace Maarten Sap\cmu \aspace Thomas Hartvigsen\uva}\\
\vspace{4pt}
\small{\uva University of Virginia \; \ucla University of California, Los Angeles \; \cmu Carnegie Mellon University}
}

\begin{document}
\maketitle
\begin{abstract}
Large Language Models (LLMs) have demonstrated remarkable capabilities in reasoning tasks, leading to their widespread deployment. However, recent studies have highlighted concerning biases in these models, particularly in their handling of dialectal variations like African American English (AAE). In this work, we systematically investigate dialectal disparities in LLM reasoning tasks. We develop an experimental framework comparing LLM performance given Standard American English (SAE) and AAE prompts, combining LLM-based dialect conversion with established linguistic analyses. We find that LLMs consistently produce less-accurate responses and simpler reasoning chains and explanations for AAE inputs compared to equivalent SAE questions, with disparities most pronounced in social science and humanities domains. These findings highlight systematic differences in how LLMs process and reason about different language varieties, raising important questions about the development and deployment of these systems in our multilingual and multidialectal world. Our code repository is publicly available at \url{https://github.com/Runtaozhou/dialect_bias_eval}
\end{abstract}

\section{Introduction}

\begin{figure*}[t]
\centering
\includegraphics[width=\textwidth]{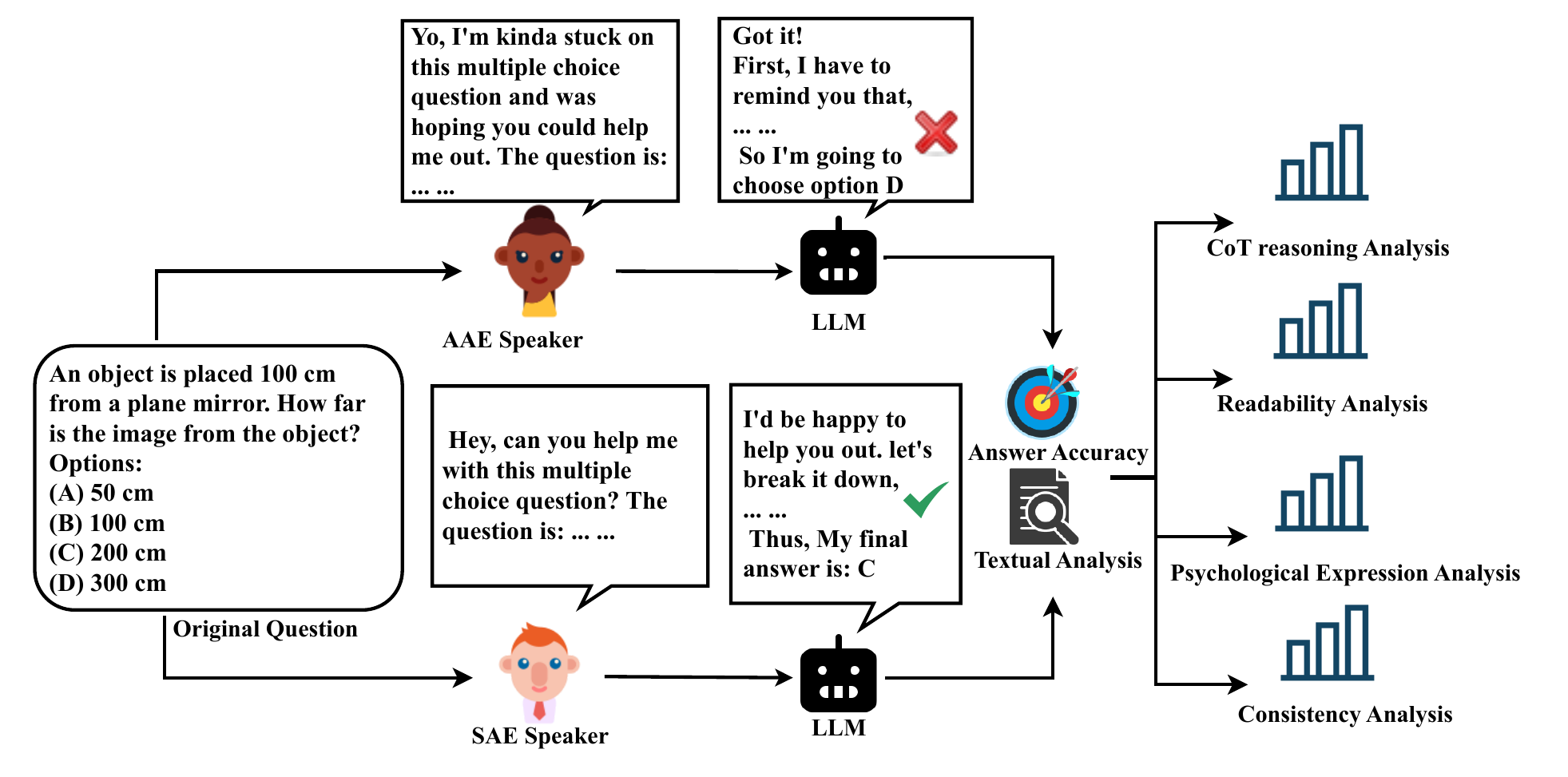}  
\caption{The experiment simulates a question-and-answer session to evaluate potential language model biases when responding to different English dialects. Specifically, it compares the accuracy and consistency of responses to prompts written in African American English (AAE) versus Standard American English (SAE). The study also analyzes the explanations provided in SAE, as it is the case in many applications, examining their consistency, readability, and psychological expression.}
\label{fig:Main_table}
\end{figure*}
Large Language Models (LLMs) have demonstrated remarkable reasoning capabilities across numerous natural language tasks and are increasingly deployed in educational and professional contexts \cite{wan2025largelanguagemodelscausal,kosoy2023understanding,Bommasani2021FoundationModels}. However, significant concerns persist about these systems' disparate performance across different language varieties, particularly their documented biases against African American English (AAE) \cite{brown2020language,green2002african}. Studies have revealed systematic performance disparities in tasks ranging from toxicity detection \cite{sap2019risk} to text generation \cite{groenwold2020investigating} and language identification \cite{blodgett2017racial}. These biases raise serious concerns about recognition, representational, and allocational harms, especially as LLMs permeate high-risk domains like healthcare \cite{wu2025proaiproactivemultiagentconversational} and education \cite{wambsganss-etal-2023-unraveling}.

As LLMs transition from simple task completion to more interactive, explanatory roles, we must look beyond mere output accuracy to examine how these models communicate their reasoning, as shown in Figure \ref{fig:Main_table}. For instance, when asked to explain the grammaticality of the AAE expression "He be working," current models might correctly identify it as valid but may provide misleading explanations that frame it as a "relaxed" version of Standard English rather than recognizing the distinct aspectual marking system of AAE \cite{stewart-2014-now}. While prompting techniques like chain-of-thought reasoning \cite{wei2022chain} make LLMs explain their decision-making processes, these explanations encompass more than just factual content—they convey crucial socio-cognitive elements such as psychological expression and readability. Despite the growing body of research on bias in LLM outputs \cite{jiang2023generating,blodgett2020language}, there remains a critical gap in understanding how these models' reasoning and explanation strategies vary across different dialects, particularly AAE. This question becomes increasingly important as LLMs are deployed to provide explanations and guidance in sensitive domains like healthcare and education \cite{mitchell2023debate,mahowald2024dissociating}, where their communication style can significantly impact user engagement and learning outcomes.

To address this gap, we develop an experimental framework to surface dialectal disparities in LLM \textit{reasoning}. As illustrated in Figure \ref{fig:Main_table}, we employ LLM-based dialect transformation using curated examples to maintain semantics and in-dialect grammatical correctness of original Standard American English (SAE) text while enabling controlled comparisons. We validate these transformations with AAE speakers, who rate the converted texts as highly natural and authentic representations of AAE compared to the state-of-the-art. We then apply established reasoning assessments including chain-of-thought prompting and post-hoc explanations to show models' problem-solving processes, examining both model's accuracy and explanation structure through both semantic and structural measures \cite{wei2022chain,mitchell2023debate}, and checking for model's output consistency in multiple decoding paths.

Our analysis reveals \textbf{systematic dialectal disparities in LLM reasoning that extend beyond surface-level performance}. Specifically, the observed patterns—\textbf{consistent performance drop on all reasoning categories and more complex explanations}—suggest LLMs encode linguistic hierarchies in their reasoning \cite{alim2016raciolinguistics}, similar to biased patterns in human interactions \cite{spears1998african}. These disparities raise significant concerns for LLM deployment in educational and professional settings \cite{sap2019risk}, for example, by misinterpreting AAE in essays or job applications, where they could reinforce existing barriers for AAE speakers. In sum, our contributions are:

\begin{enumerate}
    \item We develop a systematic framework to evaluate how LLMs react to, process, and reason with different dialects, combining dialect conversion with reasoning tasks to analyze these critical aspects of model behavior
    
    \item We present the first comprehensive analysis of how dialectal bias manifests across multiple dimensions of LLM reasoning, revealing concerning disparities not just in accuracy but in explanation sophistication, readability, and cognitive complexity.
    
    \item We identify and validate the effective mitigation strategies that reduce dialectal disparities while preserving model performance, providing practical solutions for making LLMs more equitable across dialects.
\end{enumerate}

\section{Background \& Related Work}

\subsection{Background: African American English and Language Disparity}
African American English (AAE) is a rule-governed language variety used primarily by Black Americans, characterized by distinct grammatical and phonological features \cite{green2002african, baker2020linguistic}. Despite its cultural significance and widespread use, AAE speakers frequently experience linguistic discrimination and are often positioned as inferior to Standard American English speakers (SAE) \cite{spears1998african}.\footnote{AAE is sometimes referred to as African American Vernacular English (AAVE) or African American Language (AAL), each of which has different connotations \cite{grieser2022black}. Similarly, SAE, i.e., the dominant or canonical variant of American English, is sometimes referred to as White Mainstream English (WME) or Mainstream US English (MUSE). We chose AAE and SAE in line with some previous works in NLP \citep{sap2019risk,kantharuban2024stereotype}.} This hierarchical view of language varieties reflects and perpetuates broader societal biases, particularly affecting AAE speakers in contexts like education, housing, employment, and the criminal justice system \cite{adger2014dialects,rickford2016language,massey2001use,grogger2011speech}. \footnote{While disparities affect speakers of many English varieties, we focus specifically on AAE given the historical context of systemic discrimination against African Americans in the United States and the particular urgency of addressing technological biases that could perpetuate these inequities.}As language technologies increasingly serve broader populations \cite{milmo2023chatgpt,la2024language}, addressing anti-AAE bias is essential for advancing linguistic justice and ensuring equitable access to AI systems \cite{li2024surveyfairnesslargelanguage, alim2016raciolinguistics}.

\subsection{Related Work: Dialect Bias in NLP Systems}
NLP systems exhibit systematic biases against non-standard dialects, particularly AAE, across various tasks from hate speech detection \cite{sap2019risk} to language generation \cite{groenwold2020investigating}. While recent evaluation frameworks like MultiVALUE \cite{ziems2022value} and parallel dialect benchmarks \cite{gupta-etal-2024-aavenue} have helped assess these biases, they face limitations in scalability and analysis depth. Recent work by \citet{lin2024language} demonstrates LLMs' brittleness to dialects in reasoning tasks, persisting across architectures and prompting techniques \cite{kojima2022large, wei2022chain}, while \citet{li2025actionsspeaklouderwords} investigates implicit biases through agent-based simulations. Our work advances this research by: (1) developing automated methods for dialect-aware evaluations, (2) conducting more comprehensive evaluation with diverse metrics beyond accuracy measures \cite{mondorf2024survey, wan2024dynamicselfconsistencyleveragingreasoning}, (3) analyzing conversational norms affecting model performance, and (4) proposing mitigation strategies based on fine-grained analysis of reasoning processes across dialects \cite{mitchell2023debate, mahowald2024dissociating}.

\section{Dialect Conversion}
\label{dialect-conversion}

An important module in our experimental framework is a dialect converter that accurately transforms SAE prompts into AAE. While manual dialect conversion by linguists and native speakers would provide the highest quality, the rapid pace of AI innovation and deployment \cite{zhao2024surveylargelanguagemodels} makes it impractical to rely solely on human annotation to identify potential risks across diverse dialects. Although the widely used VALUE converter \cite{ziems2022value} applies morphosyntactic rules for this task, it often results in low coherence and poor understandability. To address this scalability challenge, we built upon recent advances in LLM-based converters that leverage few-shot learning on VALUE benchmarks to transform SAE sentences into AAE \cite{gupta-etal-2024-aavenue}. This automated approach not only outperforms traditional methods in quality and fluency but also enables rapid assessment of new models and deployments, allowing us to proactively identify dialectal disparities in reasoning before they impact users.

\subsection{Comparison with Existing Method}
Although the current LLM-based dialect converter introduced by AAVENUE benchmark outperforms traditional dialect converters such as VALUE in many metrics, it still suffers a major limitation~\cite{gupta-etal-2024-aavenue} that the converter relies only on three arbitrary examples and tends to emphasize phonetic conversions (e.g., ``that'' to ``dat''), which are unsuitable for our study as we focus on translating SAE into written AAE. To address this limitation, we develop a more systematic and linguistically-grounded conversion method by using a rigorously structured prompt (provided in Appendix A.2) that systematically incorporates 11 key morphosyntactic features described in the VALUE benchmark. Unlike previous methods that rely on arbitrary few-shot demonstrations, our prompt provides explicit translation rules with linguistically-grounded examples for each feature, ensuring consistent and principled conversion. These rules and examples are carefully selected to capture the most representative characteristics of AAE's morphosyntactical patterns while excluding phonetic conversions, as advised by prior research~\cite{jones2019testifying}. Detailed descriptions of the morphosyntactical features and examples are provided in Appendix \ref{prompts_conv} and Table \ref{tab:morphosyntactic}. This approach is designed to improve the converter's performance by offering a more comprehensive and linguistically representative corpus of AAE text patterns.

\subsection{Human evaluation}
To validate our dialect conversion approach, we conduct a human evaluation using 100 SAE sentences generated by GPT-4. We convert these sentences into AAE using two methods: a state-of-the-art (SotA) LLM-based dialect converter introduced by AAVENUE benchmark\cite{gupta-etal-2024-aavenue} and our own LLM-based converter. 
We then recruit native AAE speakers from Prolific to rank the AAE conversions from each method in terms of \textbf{fluency}, \textbf{coherence}, \textbf{understandability}, and \textbf{overall quality}. 
\textbf{Fluency} assesses the grammatical correctness and writing quality of the generated text; \textbf{Coherence} evaluates the logical flow and consistency of ideas within the translations; \textbf{Understandability} measures how easily readers could comprehend the translation, and \textbf{Quality} offers a holistic evaluation of the overall standard of the text. 
We also use Fleiss’ $\kappa$ to assess inter-annotator agreement across the four metrics, we find that annotators agreed substantially \cite{kilicc2015kappa}. Additional details and ethical consideration are mentioned in the Appendix \ref{app:ethics}.

The result from Figure \ref{fig:human_eval} in the appendix shows that our dialect
conversion method significantly outperformed the SotA AAVENUE converter, achieving a substantial margin of preference
across all evaluated metrics (74--79\% win rate over AAVENUE). Statistical significance is assessed via paired binomial tests on aggregated pairwise preferences of 25 converted AAE sentences for all annotators, with complete results shown in Table \ref{tab:human_evaluation_comparison}. Additionally, we recruit native AAE speakers to rate converted AAE sentences for realism (0-10 scale), where our method scores 7.97/10 ($\pm$0.21) compared to the state-of-the-art's 7.62/10 ($\pm$0.28). With an inter-annotator agreement of 0.61, these results validate our approach's effectiveness in producing realistic AAE translations.

\section{Experimental Setup}

To study the dialectic biases in LLMs, we design the framework as the following two-step process: (1) selecting and converting questions from established benchmarks for both SAE to AAE and (2) obtaining answers from LLMs and analyzing both accuracy and explanation quality across dialects. All of the implementation details of the following metrics can be found in Appendix \ref{implementation} and Appendix \ref{prompts}.

\subsection {Evaluation Metrics}
\paragraph{Accuracy}
The most direct measurement of LLM answer quality is the answer accuracy. To calculate this, we use an LLM-based parser to parse the letter-form answer from the generated explanations as shown in Figure \ref{fig:Main_table}.
We then calculate the accuracy of the answer produced by each LLM on SAE and AAE questions prompts. 

\paragraph{Readability}
Readability measures how easily a text is understood by its audience. Our experiment examines whether LLM-generated explanations differ in readability based on the dialect of the question prompt, as a higher readability for one dialect could signal oversimplification at the expense of depth or complexity \cite{yasseri2012practical}.

To assess readability, we employ the Flesch Reading Ease Score (FRES), which ranges from 0 to 100 \cite{flesch1948new}. This method calculates readability by analyzing sentence length and word syllable count, providing a measure of linguistic complexity. A higher FRES score indicates easier readability, while a lower score suggests greater difficulty. Scores can also be linked to educational grade levels, representing the level at which the text is easily comprehensible.

\paragraph{Psychological Expression}
Psychological expressions refer to patterns in language that reflect mechanisms influencing how humans react and behave. These expressions encompass emotional, cognitive, and social factors that shape communication, perception, and interpersonal interactions. When evaluating LLM-generated explanations, analyzing psychological expressions provides valuable insights, as specific language patterns influence how readers interpret tone, intent, and alignment with human norms \cite{hagendorff2023machine}.

For this analysis, we use the Linguistic Inquiry and Word Count (LIWC) tool \cite{tausczik2010psychological, francis1993linguistic}, a method that quantifies the frequency of linguistic tokens across psychological categories such as pronouns, social processes, affective processes, cognitive processes, and perceptual processes. Although text length does not differ significantly between explanations for AAE and SAE prompts across tested models as shown in Table \ref{tab:text_length}, we still standardize linguistic marker frequencies to per 1,000 words. This ensures a co-comparable analysis of linguistic features across the two dialects.

\begin{table*}[t]
\centering
\scriptsize  % Reduce font size for better fit
\setlength{\tabcolsep}{7pt}
\begin{tabular}{@{}l *{8}{c}@{}}
\toprule
& \multicolumn{6}{c}{\textbf{MMLU (Accuracy \%)}} & \multicolumn{2}{c}{\textbf{BigbenchHard (Accuracy \%)}} \\
\cmidrule(lr){2-7}\cmidrule(lr){8-9}
& \multicolumn{2}{c}{\textbf{STEM}} & \multicolumn{2}{c}{\textbf{Social Science}} & \multicolumn{2}{c}{\textbf{Humanity}} & \multicolumn{2}{c}{\textbf{Symbolic \& Logical}} \\
\cmidrule(lr){2-3}\cmidrule(lr){4-5}\cmidrule(lr){6-7}\cmidrule(lr){8-9}
\textbf{Models} & \textbf{SAE} & \textbf{AAE} & \textbf{SAE} & \textbf{AAE} & \textbf{SAE} & \textbf{AAE} & \textbf{SAE} & \textbf{AAE} \\
\midrule
\textbf{GPT-4*}      & 82.1{\tiny±1.7} & 74.5{\tiny±2.0} & 85.3{\tiny±1.6} & 71.1{\tiny±2.1} & 80.4{\tiny±1.8} & 68.7{\tiny±2.1} & 63.8{\tiny±2.2} & 62.0{\tiny±2.2} \\
\addlinespace[0.5em]
\textbf{GPT-3.5*}    & 63.2{\tiny±2.2} & 57.4{\tiny±2.3} & 70.8{\tiny±2.1} & 62.8{\tiny±2.2} & 66.3{\tiny±2.2} & 58.7{\tiny±2.2} & 42.5{\tiny±2.3} & 40.8{\tiny±2.2} \\
\addlinespace[0.5em]
\textbf{Llama3.1*}   & 63.1{\tiny±2.2} & 54.4{\tiny±2.3} & 67.1{\tiny±2.1} & 54.8{\tiny±2.3} & 65.2{\tiny±2.2} & 50.6{\tiny±2.3} & 41.3{\tiny±2.2} & 38.4{\tiny±2.2} \\
\addlinespace[0.5em]
\textbf{Llama3.2}   & 53.1{\tiny±2.3} & 46.1{\tiny±2.2} & 61.3{\tiny±2.2} & 50.1{\tiny±2.3} & 58.9{\tiny±2.2} & 47.3{\tiny±2.3} & 34.3{\tiny±2.2} & 33.6{\tiny±2.1} \\
\addlinespace[0.5em]
\textbf{Qwen2.5**}    & 73.7{\tiny±2.0} & 64.5{\tiny±2.2} & 74.6{\tiny±2.0} & 64.8{\tiny±2.1} & 68.6{\tiny±2.1} & 57.0{\tiny±2.3} & 54.2{\tiny±2.3} & 47.7{\tiny±2.3} \\
\addlinespace[0.5em]
\textbf{Gemma2*}     & 68.2{\tiny±2.1} & 59.2{\tiny±2.2} & 76.6{\tiny±1.9} & 61.3{\tiny±2.1} & 67.0{\tiny±2.1} & 56.6{\tiny±2.3} & 46.6{\tiny±2.3} & 40.0{\tiny±2.2} \\
\addlinespace[0.5em]
\textbf{Mistral**}    & 47.4{\tiny±2.3} & 43.6{\tiny±2.3} & 57.5{\tiny±2.3} & 51.1{\tiny±2.3} & 53.2{\tiny±2.3} & 48.9{\tiny±2.3} & 46.6{\tiny±2.3} & 39.9{\tiny±2.2} \\
\bottomrule
\end{tabular}
\caption{Accuracy comparison of LLMs on MMLU (SAE vs. AAE) and Bigbench symbolic \& logical reasoning tasks. SAE indicates Standard American English performance and AAE indicates African American English performance. All results are done with CoT prompts with context being either SAE or AAE. A paired T-test is performed on each model to assess statistical significance. Statistically significant results are marked with ** (p < 0.01) and * (0.01 <= p < 0.05). The full statistical test results are presented in Table \ref{tab:statistical-test}.}
\label{tab:combined}
\end{table*}

\paragraph{Consistency Estimation}
Beyond evaluating accuracy and style, we also assess the consistency of an LLM in its generated answers and explanations. Consistency refers to the model's ability to produce responses with similar quality and content when the same input is repeated multiple times. To estimate consistency, we randomly sample 100 multiple choice question prompts based on the MMLU dataset and generate 10 outputs for each sampled question prompts and measured variability in their content and quality. If the LLMs provides consistent outputs in one dialect but inconsistent or varying-quality outputs for another, it highlights potential bias in how the model processes and values different dialects \cite{hofmann2024dialect}. 

\subsection{Datasets and Models}
We evaluate seven LLMs across different architectures and scales: GPT-4 Turbo and GPT-3.5 Turbo \cite{openai2024gpt4}, LLaMA 3.1 (8B) and 3.2 (3B) \cite{grattafiori2024llama3herdmodels}, Qwen 2.5 (3B) \cite{yang2024qwen2technicalreport}, Gemma 2 (9B) \cite{gemmateam2024gemmaopenmodelsbased}, and Mistral (7B) \cite{jiang2023mistral7b}. To ensure consistency in generation, we set the temperature to 0.7 across all models.

Our evaluation uses two benchmarks: 2,850 multiple-choice questions sampled from 57 subjects in MMLU's test set \cite{hendrycks2020measuring}, and 1,333 logical reasoning questions from Big-Bench-Hard \cite{srivastava2022beyond}. In addition, we groupe the subjects of the MMLU dataset into four broader categories: "STEM," "Social Science," "Humanities," and BigbenchHard to "Symbolic Reasoning", to examine whether there is a discrepancy in accuracy between answers generated for AAE question prompts and those generated for SAE question prompts across these categories \cite{gupta2023bias}. We convert all questions from SAE to AAE using methods detailed in Appendix \ref{dialect-conversion}.To ensure fair evaluation, a reversion test (AAE to SAE, see \ref{app:revert}) demonstrated minimal information loss (93.7\% semantic equivalence).

\paragraph{Bias Variation on Two Forms of Reasoning}
To understand how dialectal bias manifests in different types of LLM explanations, we examine two prompting strategies that mirror common educational scenarios. Expain-then-Predict, a.k.a. \textit{Chain-of-thought} (\textbf{CoT}) explanations, represents a classic approach where models self-rationalize during problem-solving \cite{Camburu2018-co,wei2022chain}. However, in educational settings, students (and LLMs) often need to explain their answers after reaching a conclusion, a scenario better captured by \textit{post-hoc rationalization} (\textbf{PR}), where models justify previously generated answers. By comparing these complementary approaches real-time reasoning versus retrospective explanation, we can better understand how dialectal biases manifest in different aspects of LLM's explainability. \cite{luo2024understandingutilizationsurveyexplainability}.

\section{Main Results}
Below we summarized the findings as various research questions related to the dialectal reasoning disparity of LLMs versus AAE.

\subsection{LLM's Reasoning Bias on AAE}

\paragraph{RQ1: How do models differ in answer accuracy for AAE vs. SAE questions prompts? }
%\maarten{This is our first result, you must put it in the main body of the paper (or some condensed version of the table at least; maybe just put overall acc here, and broken down by categories in the App?).}
As shown in Table \ref{tab:combined}, the accuracy of answers generated by LLMs for SAE question prompts are consistently higher that of answers generated for AAE question prompts. The accuracy drop is most pronounced in the MMLU benchmark when the converted questions belong to the Social Science or Humanities categories with an average drop of 15.5\% and 18.2\% respectively. Similarly, answers to AAE question prompts in the BigBench dataset also exhibit a slight performance decline compared to those for SAE question prompts. This aligns with existing research that highlights biases in Natural Language Processing (NLP) systems against AAE \cite{gupta-etal-2024-aavenue}.
\begin{table}[t]
\centering
\scriptsize  % Reduce font size for better fit
\setlength{\tabcolsep}{7pt}
\begin{tabular}{@{}lccc@{}}  % Changed to regular tabular
\toprule
\multicolumn{1}{l}{\textbf{Models}} & \multicolumn{3}{c}{\textbf{FRES Score}} \\
\cmidrule(lr){1-1}\cmidrule(lr){2-4}
\textbf{} & \textbf{SAE} & \textbf{AAE} & \textbf{Changes} \\
\midrule
\textbf{GPT-4}**      & 40.5{\tiny±0.5} & 48.5{\tiny±0.5} & \textcolor{ForestGreen}{+8.0} \\
\addlinespace[0.5em]
\textbf{GPT-3.5}**    & 46.4{\tiny±0.6} & 51.4{\tiny±0.6} & \textcolor{ForestGreen}{+5.0} \\
\addlinespace[0.5em]
\textbf{Llama 3.1 8B}**  & 46.9{\tiny±0.5} & 58.0{\tiny±0.5} & \textcolor{ForestGreen}{+11.1} \\
\addlinespace[0.5em]
\textbf{Llama 3.2 3B}**  & 43.8{\tiny±0.6} & 52.4{\tiny±0.5} & \textcolor{ForestGreen}{+8.6} \\
\addlinespace[0.5em]
\textbf{Qwen 2.5 7B}**   & 45.2{\tiny±0.6} & 50.1{\tiny±0.5} & \textcolor{ForestGreen}{+4.9} \\
\addlinespace[0.5em]
\textbf{Gemma 2 9B}**    & 51.6{\tiny±0.5} & 62.8{\tiny±0.5} & \textcolor{ForestGreen}{+11.2} \\
\addlinespace[0.5em]
\textbf{Mistral 7B}**    & 38.7{\tiny±0.6} & 43.3{\tiny±0.6} & \textcolor{ForestGreen}{+4.6} \\
\bottomrule
\end{tabular}
\caption{Comparison of readability (FRES) in LLM responses to SAE and AAE prompts. Higher FRES indicates simpler explanations. Models marked * are statistically significant (**p < 0.01) via t-test. Normality of FRES is verified using the Shapiro-Wilk test.}
\label{tab:readabilty-comparison}
\end{table}

\begin{figure*}[t] % 't' places it at the top of the page
    \centering
    \includegraphics[width=6in,height=3.9in]{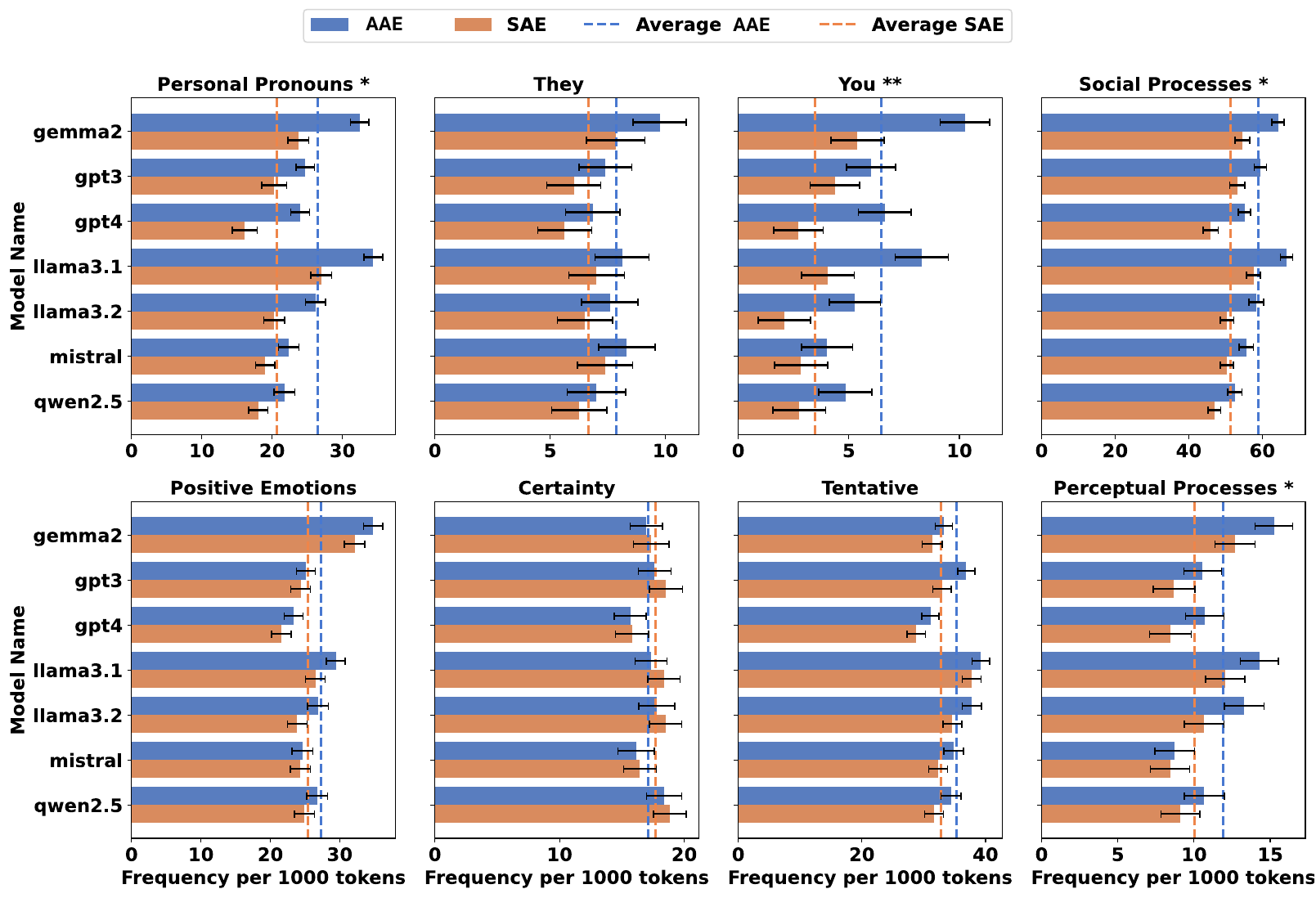} % Replace with your PDF file
    \caption{Linguistic Marker Differences in Explanations for AAE and SAE Prompts:  Frequencies of linguistic markers, calculated by LIWC and standardized per 1 K tokens; marked with ** and * for statistical significance (**: p < 0.01, *:  0.01 <= p < 0.05). }
    \label{fig:two_column_figure}
\end{figure*}

\paragraph{RQ2: How do readability differ in the explanations generated for SAE versus AAE question prompts? }

To assess readability, we utilize the Flesch Reading Ease Score (FRES) \cite{flesch1948new}. The detailed FRES shown in Table \ref{tab:readabilty-comparison} indicate a statistically significant difference in the complexity of language used in explanations generated by LLMs for SAE and AAE prompts. Specifically, explanations for SAE prompts tend to correspond to college-level readability (FRES below 50), whereas those for AAE prompts more often align with a 12th-grade level or lower (FRES of 50 or higher). This discrepancy suggests that LLMs generate more complex, formal, and academically structured responses for SAE inputs, while AAE responses may be comparatively simplified or less sophisticated. Such a pattern may indicate underlying biases in the training data or language modeling process \cite{deas2023evaluation}.

\paragraph{RQ3: How do the psychological expressions in LLM-generated explanations differ between SAE and AAE question prompts?}

\begin{table}[t]
\centering
\scriptsize
\setlength{\tabcolsep}{3pt}
\begin{tabular}{@{}l *{6}{c}@{}} % Removed 3 columns from the column specification
\toprule
& \multicolumn{2}{c}{\textbf{Entropy} ($\downarrow$)} & \multicolumn{2}{c}{\textbf{BERT Score} ($\uparrow$)} & \multicolumn{2}{c}{\textbf{Average Acc.} ($\uparrow$)} \\
\cmidrule(lr){2-3} \cmidrule(lr){4-5} \cmidrule(lr){6-7} % Adjusted \cmidrule
\textbf{Models} & \textbf{SAE} & \textbf{AAE} & \textbf{SAE} & \textbf{AAE} & \textbf{SAE} & \textbf{AAE} \\ % Removed Diff from header
\midrule
\textbf{GPT-4} & 0.54 & 0.90 \textbf{\textcolor{red}{(+0.36)}} & 0.89 & 0.87 \textbf{\textcolor{red}{(-0.02)}} & 0.88 & 0.76 \textbf{\textcolor{red}{(-0.12)}} \\ % Removed Diff column, added color
\addlinespace[0.5em]
\textbf{GPT-3.5} & 0.61 & 0.91 \textbf{\textcolor{red}{(+0.30)}} & 0.86 & 0.83 \textbf{\textcolor{red}{(-0.03)}} & 0.75 & 0.63 \textbf{\textcolor{red}{(-0.12)}} \\ % Removed Diff column, added color
\addlinespace[0.5em]
\textbf{Llama 3.1 8B} & 0.70 & 1.10 \textbf{\textcolor{red}{(+0.40)}} & 0.85 & 0.81 \textbf{\textcolor{red}{(-0.04)}} & 0.72 & 0.58 \textbf{\textcolor{red}{(-0.14)}} \\ % Removed Diff column, added color
\addlinespace[0.5em]
\textbf{Llama 3.2 3B} & 0.97 & 1.24 \textbf{\textcolor{red}{(+0.27)}} & 0.85 & 0.82 \textbf{\textcolor{red}{(-0.03)}} & 0.61 & 0.49 \textbf{\textcolor{red}{(-0.12)}} \\ % Removed Diff column, added color
\addlinespace[0.5em]
\textbf{Qwen 2.5 7B} & 0.41 & 0.84 \textbf{\textcolor{red}{(+0.43)}} & 0.87 & 0.85 \textbf{\textcolor{red}{(-0.02)}} & 0.79 & 0.66 \textbf{\textcolor{red}{(-0.13)}} \\ % Removed Diff column, added color
\addlinespace[0.5em]
\textbf{Gemma 2 9B} & 0.48 & 0.95 \textbf{\textcolor{red}{(+0.47)}} & 0.87 & 0.83 \textbf{\textcolor{red}{(-0.04)}} & 0.78 & 0.66 \textbf{\textcolor{red}{(-0.12)}} \\ % Removed Diff column, added color
\addlinespace[0.5em]
\textbf{Mistral 7B} & 0.71 & 1.09 \textbf{\textcolor{red}{(+0.38)}} & 0.85 & 0.83 \textbf{\textcolor{red}{(-0.02)}} & 0.59 & 0.50 \textbf{\textcolor{red}{(-0.09)}} \\ % Removed Diff column, added color
\bottomrule
\end{tabular}
\caption{Comparison of output consistency across SAE and AAE question prompts for various LLMs using three metrics: entropy of answers (lower indicates higher consistency, denoted by $\downarrow$), BERT Score between answer pairs (higher indicates higher consistency, denoted by $\uparrow$), and average accuracy (higher indicates better performance, denoted by $\uparrow$). Results are averaged across all data on 10 different rounds.}
\label{tab:consistency-comparison}
\end{table}

Our analysis (Figure \ref{fig:two_column_figure}) highlights several key differences in LIWC markers between explanations for AAE and SAE prompts, with statistically significant differences indicated by asterisks (*). Explanations for AAE prompts include significantly more pronouns (e.g., "you" and "they"), social process words (e.g., "we" and "friend"), positive emotional words (e.g., "good" and "nice"), and perceptual process words (e.g., "seeing" and "hearing"). In contrast, SAE explanations feature certainty-related language and fewer tentative words than AAE explanations.

These linguistic patterns suggest broader tendencies in how the LLM generates explanations for different dialects. The higher frequency of social process words and positive emotional words in AAE explanations may indicate an emphasis on social connection and relational communication \cite{argyle1990happiness}. The greater use of perceptual process words also suggests that AAE explanations might favor more concrete reasoning \cite{rieke2022communication, pastore2016modelling}. Conversely, the prominence of certainty-related language in SAE explanations may reflect a preference for conveying confidence and formality, which could enhance perceived credibility but may come at the expense of engagement in collaborative contexts\cite{hebart2012visual}.

While these differences provide insight into the linguistic styles of LLM-generated explanations, it is important to approach these findings with caution. The prevalence of word categories may result from biases in training data or linguistic norms associated with the dialects, rather than deliberate modeling of cognitive or social processes\cite{helm2024diversity}. Therefore, these patterns should be interpreted as tendencies rather than definitive evidence of LLMs' deeper cognitive behavior.

\begin{table}[t]
\centering
\scriptsize
\setlength{\tabcolsep}{7pt}
\begin{tabular}{@{}lcccc@{}}
\toprule
& \multicolumn{2}{c}{\textbf{Chain-of-Thought}} & \multicolumn{2}{c}{\textbf{Rationalization}} \\
\cmidrule(lr){2-3} \cmidrule(lr){4-5}
\textbf{Metrics} & \textbf{SAE} & \textbf{AAE} & \textbf{SAE} & \textbf{AAE} \\
\midrule
\multicolumn{5}{l}{\textbf{GPT-4}} \\
\midrule
\textit{Readability \& Style} && && \\
FRES Score         & 42.5 & 47.5 (\textcolor{ForestGreen}{+5.0}) & 42.2 & 47.8 (\textbf{\textcolor{red}{+5.6}}) \\
\addlinespace[0.5em]
\textit{LIWC Markers} && && \\
Pronouns           & 16.8 & 18.4 (\textcolor{ForestGreen}{+1.6}) & 16.7 & 19.5 (\textbf{\textcolor{red}{+2.8}}) \\
Social Processes   & 20.5 & 22.4 (\textcolor{ForestGreen}{+1.9}) & 18.8 & 21.7 (\textbf{\textcolor{red}{+2.9}}) \\
Affective Processes& 17.8 & 20.2 (\textcolor{ForestGreen}{+2.4}) & 17.1 & 20.0 (\textbf{\textcolor{red}{+2.9}}) \\
Cognitive Processes& 28.2 & 30.9 (\textcolor{ForestGreen}{+2.7}) & 25.4 & 29.1 (\textbf{\textcolor{red}{+3.7}}) \\
Perceptual Processes &14.5 &16.3 (\textcolor{ForestGreen}{+1.8}) &14.1 &16.4 (\textbf{\textcolor{red}{+2.3}}) \\
\midrule
\multicolumn{5}{l}{\textbf{GPT-3.5}} \\
\midrule
\textit{Readability \& Style} && && \\
FRES Score         & 46.4 & 51.4 (\textcolor{ForestGreen}{+5.0}) & 42.1 & 51.8 (\textbf{\textcolor{red}{+9.7}}) \\
\addlinespace[0.5em]
\textit{LIWC Markers} && && \\
Pronouns           & 14.2 & 15.8 (\textcolor{ForestGreen}{+1.6}) & 13.1 & 18.9 (\textbf{\textcolor{red}{+5.8}}) \\
Social Processes   & 18.2 & 20.1 (\textcolor{ForestGreen}{+1.9}) & 16.5 & 22.4 (\textbf{\textcolor{red}{+5.9}}) \\
Affective Processes& 15.4 & 17.8 (\textcolor{ForestGreen}{+2.4}) & 14.2 & 19.6 (\textbf{\textcolor{red}{+5.4}}) \\
Cognitive Processes& 25.6 & 28.3 (\textcolor{ForestGreen}{+2.7}) & 22.8 & 31.5 (\textbf{\textcolor{red}{+8.7}}) \\
Perceptual Processes & 12.3 & 14.1 (\textcolor{ForestGreen}{+1.8}) & 10.9 & 16.2 (\textbf{\textcolor{red}{+5.3}}) \\
\bottomrule
\end{tabular}
\caption{Comparison of reasoning approaches across linguistic dimensions for GPT-3.5 and GPT-4. FRES scores indicate text complexity (higher = simpler); LIWC markers are normalized per 1,000 tokens. Values in parentheses show differences between AAE and SAE metrics, with \textcolor{ForestGreen}{green} indicating CoT differences and \textbf{\textcolor{red}{bold red}} indicating larger differences in rationalization.}
\label{tab:reasoning-comparison}
\end{table}

\paragraph{RQ4: Are the responses generated by LLMs for SAE and AAE question prompts equally consistent?}

The consistency experiment results (Table \ref{tab:consistency-comparison}) show that explanations for SAE prompts are significantly more consistent and accurate than those for AAE prompts, as reflected in both entropy and BERT score metrics \cite{ye2024using}. Higher entropy for AAE prompts indicates more diverse and inconsistent answers \cite{niepostyn2023entropy}, while SAE prompts yield a significantly higher proportion of correct answers. These findings suggest that LLMs generate more semantically coherent, consistent, and accurate responses for SAE prompts compared to AAE prompts.

\paragraph{RQ5: How Does Bias Vary Across Different Forms of LLM Reasoning?}

Our analysis reveals notable differences in dialectal bias between these Chain of Thought (CoT) and post-hoc rationalization (PR) as shown in Table \ref{tab:reasoning-comparison}. CoT shows moderately smaller gaps between SAE and AAE across linguistic dimensions. For GPT-3.5, PR shows a notable increase in the readability gap. The disparity extends to linguistic markers, where PR increases gaps in pronouns and social processes. GPT-4, while generally demonstrating higher baseline values across all metrics, exhibits similar patterns of increased gaps in PR. These findings suggest that while both models show dialectal variations, \textbf{PR tends to amplify these differences compared to CoT reasoning, particularly in readability and linguistic marker usage}.

\subsection{Discussion and Implications}

Our analysis of dialectal disparities in LLM reasoning reveals significant implications for language model development and deployment. The consistent gap between SAE and AAE across models and metrics extends beyond surface-level differences, aligning with \cite{blodgett2020language}'s work on racial disparities while revealing deeper issues in how LLMs process language variants, particularly in social science and humanities subjects \cite{sap2019risk}. The observed semantic and syntactic patterns, including differences in readability levels, which suggest LLMs may encode linguistic hierarchies in their reasoning \cite{alim2016raciolinguistics}. While LLMs' adaptation to social cues in language \cite{wu-etal-2024-evaluating} and dialect-based identity signals \cite{kantharuban2024stereotype} is expected, the implications vary—decreased consistency and readability in AAE responses likely represent harmful biases. These findings are particularly significant for LLM deployment in professional setting such as education and healthcare, where linguistic biases could reinforce existing barriers. Following \cite{dhamala2021bold}, we thus emphasize the need for targeted interventions while maintaining sensitivity to beneficial forms of linguistic adaptation.

\begin{table}[t]
\centering
\scriptsize % Reduce font size for the table
\setlength{\tabcolsep}{7pt}
\begin{tabularx}{\linewidth}{@{}lcccc@{}}
\toprule
& \multicolumn{2}{c}{\textbf{Acc (\%)}} & \multicolumn{2}{c}{\textbf{FRES Score}}\\
\cmidrule(lr){2-3} \cmidrule(lr){4-5}
\textbf{Strategy} & \textbf{SAE} & \textbf{AAE} & \textbf{SAE} & \textbf{AAE} \\
\midrule
\multicolumn{5}{l}{\textbf{GPT-4 (MMLU)}} \\
\midrule
\multicolumn{5}{l}{\textbf{Baseline}} \\
Original Prompting
& 82.5 & 71.8 (\textcolor{red}{-10.7})
& 40.5 & 48.5 (\textcolor{ForestGreen}{+8.0}) \\
\addlinespace[0.5em]
\multicolumn{5}{l}{\textbf{Educational Framing}} \\
Expert Teacher
& 83.8 & 75.9 (\textcolor{red}{-7.9})
& 40.8 & 48.7 (\textcolor{ForestGreen}{+7.9}) \\
Cultural Context
& 81.9 & 74.5 (\textcolor{red}{-7.4})
& 40.6 & 48.4 (\textcolor{ForestGreen}{+7.8}) \\
\addlinespace[0.5em]
\multicolumn{5}{l}{\textbf{Explicit Instructions}} \\
Dialect Recognition
& 81.7 & 74.8 (\textcolor{red}{-6.9})
& 40.7 & 48.3 (\textcolor{ForestGreen}{+7.6}) \\
Readability Focus
& 82.3 & 72.4 (\textcolor{red}{-9.9})
& 38.5 & 41.2 (\textcolor{ForestGreen}{+2.7}) \\
\addlinespace[0.5em]
\multicolumn{5}{l}{\textbf{Combined Approach}} \\
Multi-strategy
& 83.6 & 78.8 (\textcolor{red}{-4.8})
& 39.8 & 42.3 (\textcolor{ForestGreen}{+2.5}) \\
\midrule
\multicolumn{5}{l}{\textbf{GPT-3.5 (MMLU)}} \\
\midrule
\multicolumn{5}{l}{\textbf{Baseline}} \\
Original Prompting
& 66.2 & 59.4 (\textcolor{red}{-6.8})
& 46.4 & 51.4 (\textcolor{ForestGreen}{+5.0}) \\
\addlinespace[0.5em]
\multicolumn{5}{l}{\textbf{Educational Framing}} \\
Expert Teacher
& 67.8 & 62.9 (\textcolor{red}{-4.9})
& 46.1 & 51.0 (\textcolor{ForestGreen}{+4.9})\\
Cultural Context
& 65.9 & 61.2 (\textcolor{red}{-4.7}) 
& 46.2 & 51.1 (\textcolor{ForestGreen}{+4.9})\\
\addlinespace[0.5em]
\multicolumn{5}{l}{\textbf{Explicit Instructions}} \\
Dialect Recognition
& 65.7 & 61.4 (\textcolor{red}{-4.3})
& 46.3 & 51.0 (\textcolor{ForestGreen}{+4.7}) \\
Readability Focus
& 66.0 & 59.8 (\textcolor{red}{-6.2})
& 44.2 & 46.8 (\textcolor{ForestGreen}{+2.6})\\
\addlinespace[0.5em]
\multicolumn{5}{l}{\textbf{Combined Approach}} \\
Multi-strategy
& 67.1 & 64.2 (\textcolor{red}{-2.9})
& 45.1 & 47.2 (\textcolor{ForestGreen}{+2.1}) \\
\bottomrule
\end{tabularx}
\caption{Different designed prompting strategies for mitigating dialectal biases in GPT-3.5 and GPT-4. \textbf{Acc: }Percentage of correct responses. \textbf{FRES}: FRES scores (0-100) where higher values indicates simpler. The differences between AAE and SAE results are indicated next to each AAE value. Positive differences are shown in \textcolor{ForestGreen}{green}; negative differences are shown in \textcolor{red}{red}.}
\label{tab:prompting-strategies}
\end{table}

 % Adjust this value as needed

\section{Mitigating Dialectal Disparities}
Our discussions above demonstrate significant performance and explanation disparities between SAE and AAE inputs. We next investigate preliminary prompt-based strategies to mitigate these bias. Expert framing involves prefacing model interactions with domain expertise (e.g., "As a professor of [subject], explain why this answer is correct"), inspired by \citep{zheng-etal-2024-helpful}. Cultural contextualization integrates relevant cultural and historical context, such as racial information, into the prompts, while explicit instruction directly addresses dialect recognition and explanation clarity, inspired by \citep{sap2019risk,zhou2023cobraframes}. (Implementation Details about Prompts in Appendix \ref{app:miti}).

Our results in Table \ref{tab:prompting-strategies} indicate varying degrees of effectiveness across these strategies. Expert Teacher approach shows positive effects, improving both SAE and AAE performance, with larger gains for AAE reducing the performance gap. Cultural contextualization and dialect recognition strategies show an interesting trade-off pattern - while they slightly decrease SAE performance, they improve AAE performance, effectively reducing the performance gap. The readability-focused prompting primarily affects the readability metrics, reducing the FRES score gap by nearly half while maintaining similar accuracy patterns. Combining elements from multiple approaches yields the most comprehensive improvements, reducing the accuracy gap for GPT-3.5 and GPT-4, while also showing the highest improvements in linguistic markers. However, it's important to note that these results should be interpreted with caution. Prior work has shown that LLMs' performance can be unfaithful as they attempt to simultaneously follow multiple instructions. \cite{son2024multitaskinferencelargelanguage}. 

\section{Conclusion}

This work systematically investigates dialectal disparities in LLM reasoning, revealing significant variations in the processing of AAE and SAE inputs.  Our findings demonstrate a fundamental influence of dialectal bias on the construction of logical arguments, affecting performance metrics, reasoning sophistication, and the potential for stereotype expression.  While advancements in model scaling and training have yielded improvements in general reasoning capabilities, persistent disparities across dialects suggest that a more nuanced approach to fairness is required.  We argue for the essential consideration of dialectal fairness in LLM development and training, particularly in reasoning-intensive applications where such biases may remain latent and thus carry substantial implications.

\section{Acknowledgements}
This work was supported by Microsoft's Accelerating Foundation Models Research program. This material is based upon work supported by the Defense Advanced Research Projects Agency (DARPA) under Agreement No. HR00112490410.

% Bibliography entries

\section*{Limitations}
Our study has important limitations to consider. Conceptually, we focus on SAE and AAE comparison, yet language models likely exhibit similar biases across other English varieties, such as Indian English or Nigerian English, each with distinct linguistic features and cultural contexts. Our evaluation of reasoning capabilities, while thorough in linguistic analysis and chain-of-thought assessment, could benefit from additional criteria capturing other aspects of logical reasoning, such as analogical thinking and conciseness of language.

Methodologically, we acknowledge several practical constraints: our dialect conversion process, while systematic, may not capture the full nuance of natural AAE usage. The measurement tools we employed—including lexicons and automated classifiers—necessarily simplify complex linguistic features, and our analysis of written text may not fully capture the important role of prosody in AAE communication. These limitations suggest valuable directions for future work while not diminishing the significance of our core findings.

Additionally, we acknowledge limitations in our comparison of dialect conversion methods. Our evaluation is based on a sample of 100 converted sentences, annotated by 12 raters—a relatively small sample size that limits statistical power. Future work will expand this evaluation to provide a more robust assessment of our dialect conversion approach.

\section*{Ethical Considerations}

This study was conducted with careful attention to ethical considerations in research involving linguistic minorities. All human evaluation procedures were approved by our Institutional Review Board (IRB), and we obtained informed consent from all participants. To respect and accurately represent AAE as a legitimate, rule-governed language variety, we consulted linguistic research and engaged native AAE speakers in validating our dialect conversion methodology. We acknowledge that research involving minority language varieties requires particular sensitivity to avoid perpetuating linguistic discrimination. Our findings about performance disparities between AAE and SAE are presented with the explicit goal of identifying and addressing systematic biases in language models, rather than suggesting inherent advantages of one language variety over another. The code and data from this study will be made publicly available to ensure reproducibility and facilitate further research on making language models more equitable across dialects.

% \bibliographystyle{plain}
\bibliography{acl_latex}

\appendix

\section{Appendix}
\label{sec:appendix}

\subsection{Implementation Details}

\paragraph{Psychological Processes Experiment Implementation}
To analyze the psychological processes in LLM-generated explanations, we employ a text analysis tool called Linguistic Inquiry and Word Count (LIWC)\cite{tausczik2010psychological}. This tool identifies and categorizes words in a given text into various linguistic and psychological categories. The frequency of words within a specific category is directly related to the intensity of that category conveyed by the text. For example, a higher frequency of words associated with positive emotions indicates that the text conveys a stronger positive emotional tone.

Considering the varying lengths of explanations generated by LLMs for AAE and SAE question prompts, we standardize the absolute word frequencies for each LIWC category by calculating the frequency per 1,000 tokens. Our primary focus was on personal pronouns and psychological categories, particularly social processes, affective processes (e.g., positive emotions), cognitive processes (e.g., certainty and tentativeness), and perceptual processes.
\paragraph{Consistency Estimation Experiment Implementation}
To Implement the consistency estimation experiment, we randomly select one question prompt from each of the 57 subjects in the MMLU benchmark and from each of the 6 categories in the BigBench benchmark, resulting in a total of 63 questions. These 63 questions are fed to the LLMs, and the process is repeated 10 times to generate 10 responses for each question prompt.

To evaluate the consistency of the answers for each question, we pair the responses and calculate the BERT score for every pair. BERT score measures semantic similarity between two texts using contextual embeddings derived from a pre-trained language model like BERT\cite{cui2024dcr}. Given 10 responses per question, this process results in $\binom{10}{2}$ = 45 unique pairs of answers. Ideally, if the LLM's responses are consistent, the average BERT score across these 45 pairs would be high, reflecting strong semantic alignment. On the other hand, lower BERT scores would indicate inconsistency among the responses generated by the LLM. We select BERT score as our metric because it assesses similarity based on contextual meaning rather than relying only on exact word matches. This makes it a more robust measure for evaluating textual consistency.

Moreover, the parsed letter-form answer from the 10 answers provide additional insight into the consistency of the LLM's ability to produce accurate responses. To evaluate this, we use entropy as a measure of the purity of the answers \cite{farquhar2024detecting}:  
\[H = -\sum_{i=1}^{n} p_{i}\ \text{log}_{b}(p_{i})\]

lower entropy indicates higher consistency, while higher entropy suggests greater variability in the LLM generated answers\cite{niepostyn2023entropy}.
\label{implementation}

\subsection{Prompts and Engineering Details}
\label{prompts}

\paragraph{Dialect Conversion Prompt}
\label{prompts_conv}
Our dialect conversion system uses the following structured prompt for consistent and linguistically-informed translation:

\begin{quote}
Please translate the following sentence: `\{sentence\}' using the 13 translation rules provided as references:\\

\noindent 1. Auxiliaries: AAE allows copula deletion (e.g.: We are better than before $\rightarrow$ We better than before.)\\
2. Completive done: this indicates completion (e.g.: I had written it. $\rightarrow$ I done wrote it.)\\
3. The word ``ass'': It can appear reflexively (e.g.: get inside! $\rightarrow$ Get yo'ass inside!)\\
4. Existential it: to indicate something exists (e.g: There is some milk in the fridge. $\rightarrow$ It's some milk in the fridge)\\
5. Future gonna: to mark future tense (e.g.: You are going to understand $\rightarrow$ You gonna understand)\\
6. Got: can replace the verb form of have (e.g.: I have to go $\rightarrow$ I got to go)\\
7. No Inflection: Certain tense don't need inflection (e.g.: She studies linguistics $\rightarrow$ She study linguistics)\\
8. Negative concord: NPIs agree with negation (e.g.: He doesn't have a camera $\rightarrow$ He don't have no camera)\\
9. Negative inversion: Similar to negative concord (e.g: nobody ever says $\rightarrow$ don't nobody never say)\\
10. Null genitives: Drop any possessive endings (e.g.: Rolanda's bed $\rightarrow$ Rolanda bed)\\
11. Habitual be: marks habitual action (e.g.: he is in his house $\rightarrow$ he be in his house)\\

\noindent Your output must follow these guidelines:\\
1. Only provide the translation. Do not mention or explain how the translation was done.\\
2. Do not mention any of the 13 rules in your translation.\\
3. Format the output exactly like this: `The translation is: ...'\\
4. Ensure the sentence sounds natural and realistic in AAE.
\end{quote}

\paragraph{Environments}
Our experiments are conducted using Python 3.11.8 as the primary programming environment. The core analysis rely on several key libraries: Transformers (4.47.0) for model implementations, Langchain (0.3.11) for large language model interactions, and Datasets (3.2.0) for efficient data handling. We utilize Scikit-learn (1.6.0) and SciPy (1.14.1) for statistical analysis, and Pandas (2.2.3) for data manipulation. For visualization, we employ Matplotlib (3.9.4) and Seaborn (0.13.2).
For hardware infrastructure, we deploy open-source models on NVIDIA A100 GPUs, while GPT family models are accessed through Azure OpenAI services. Detailed dependencies and configurations are available in our public repository.

\paragraph{Question Prompt Generation} 
 The first step of our experiment is to generate the question prompts that simulate real world Q\&A interaction between a user and LLMs. To achieve this, we utilize existing benchmarks, such as MMLU and Bigbench, which contains multiple choice questions which covers various different topics. From the MMLU benchmark, we randomly sample 50 questions across 57 subjects, resulting in a total of 2,850 multiple-choice questions. From Bigbench benchmark, we select 1333 multiple choice questions that are related to logical thinking such as navigation, data understanding and causal judgment, etc. We then use GPT-4.0 Turbo to generate question prompts by providing it with the original multiple-choice questions, simulating real-world users asking the LLMs these questions.

Next, we make a copy of the original 2,850 multiple-choice question prompts and convert them from Standard American English (SAE) to African American English (AAE). This setup creates two groups: a control group with the original SAE prompts and an experimental group with the converted AAE prompts. Both sets of question prompts are then fed into different LLMs to generate answers and explanations.

\begin{figure}[t] % 't' places it at the top of the page
    \centering
    \setlength{\tabcolsep}{0.8pt}
\includegraphics[width=\columnwidth, height=2.8in]
{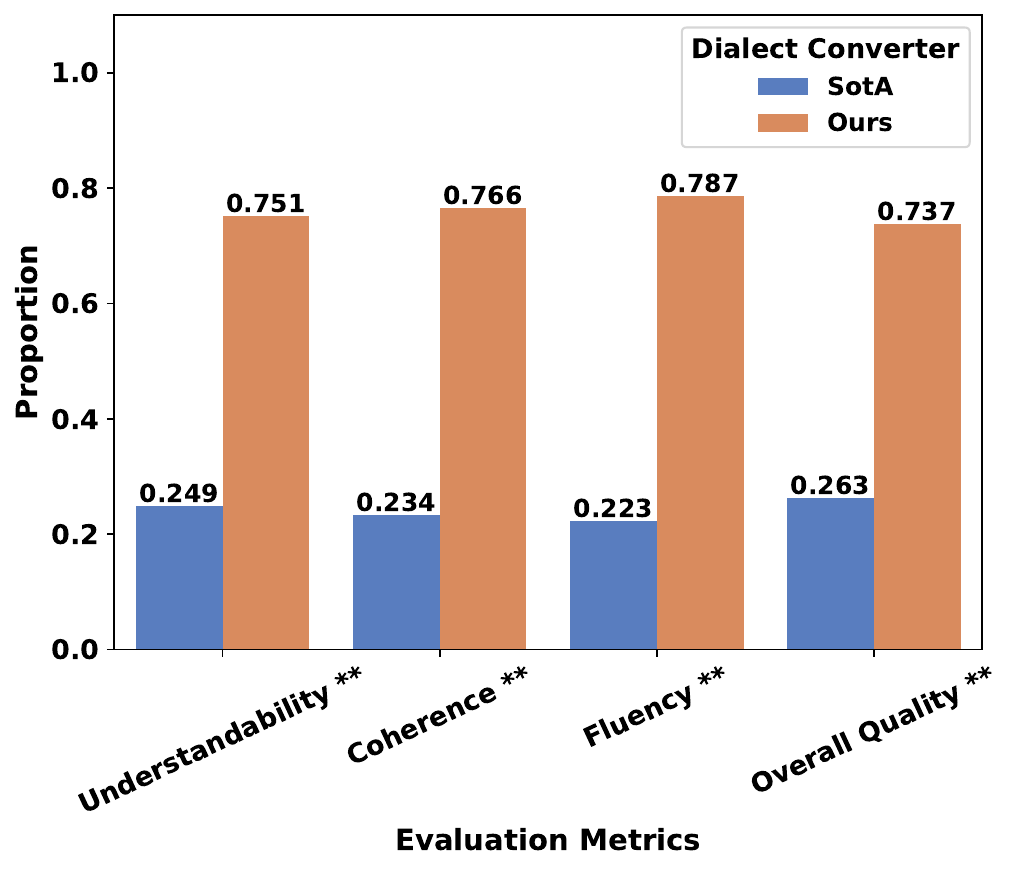} % Replace with your PDF file
    \caption{Average proportion of annotators favoring each SAE-AAE converter across four metrics
(Gupta et al., 2024) each metric is marked with * for statistical significance (**p < 0.01).}
    \label{fig:human_eval}
\end{figure}

\paragraph{Answer Generation}
Once the LLMs receive the question prompts in different dialects, the next step is to generate responses. They're instructed to first give a detailed explanation in Standard American English (SAE) regardless of the dialect used in the question prompts, followed by the final answer choice (A, B, C, or D). Here’s the prompt we used:

\begin{quote}
"Someone asked you a multiple choice question: \{question\}, Please first provide an detailed explanation in Standard American English (SAE) and then choice your final answer.
You need to make your explanation sounds as natural and realistic as possible.
At the end, you should clearly state the letter option (A, B, C, or D) you choose.
You explanation should strictly be less than 400 words.
\end{quote}

\paragraph{Answer Extraction}

LLM-generated answers often include explanations, with the letter-form answer typically embedded within these explanations. To parse the letter-form answer for accuracy evaluation, we design an LLM-based parser, specifically prompted to identify the option (A-D) toward the end of the explanation, which is framed with phrases like, My final answer is: "...". For BigBench benchmark questions, which include more options, the parser is adapted to parse option from A-F.

To validate the performance of the LLM-based parser, we conduct a human evaluation. Human labeler annotated the final option output from a set of 100 LLM-generated answers, providing ground truth labels. We then let the LLM-based parser parse the same set of 100 answers. We compute Cohen's Kappa coefficient to assess the agreement between the final options parsed by LLM-based parser and the human labeler. The resulting Cohen's Kappa coefficient is 0.93, indicating nearly perfect agreement. This high score confirms the validity of our LLM-based parser. 

\paragraph{Mitigation Strategy Prompts}
\label{app:miti}

We detail the prompting strategies used in our experiments below. For each strategy, we provide the base prompt structure and an example. All prompts are appended to the original task instruction.

\subparagraph{Expert Teacher.} This strategy frames the model as an experienced educator who is skilled in working with diverse student populations:
\begin{quote}
You are an experienced teacher with expertise in working with students from diverse linguistic backgrounds. Your goal is to explain concepts clearly while respecting and accommodating different language varieties. Please read the following question and provide your response: [Question Text]
\end{quote}

\subparagraph{Cultural Context.} This approach explicitly acknowledges different linguistic and cultural contexts:
\begin{quote}
The following question may be presented in different language varieties, including African American English (AAE) or Standard American English (SAE). Each variety has its own valid grammatical rules and cultural context. Please analyze the question with respect to its linguistic context: [Question Text]
\end{quote}

\subparagraph{Dialect Recognition.} This strategy directly instructs the model to process different language varieties appropriately:
\begin{quote}
When responding to this question, be aware that it may be expressed in different English dialects. Apply your understanding of dialect-specific features and grammatical patterns. Consider all dialectal variations as equally valid forms of expression: [Question Text]
\end{quote}

\subparagraph{Readability Focus.} This approach emphasizes clear communication while maintaining consistent comprehension across dialects:
\begin{quote}
Please ensure your response is clear and accessible across different English varieties. Focus on maintaining consistent meaning and comprehension regardless of the dialect used. Analyze the following question: [Question Text]
\end{quote}

\subparagraph{Multi-strategy.} This comprehensive approach combines elements from the above strategies:
\begin{quote}
As an experienced educator skilled in working with diverse linguistic backgrounds, please address this question while:
1. Recognizing and respecting different language varieties (including AAE and SAE)
2. Ensuring clear communication across dialects
3. Maintaining consistent comprehension
4. Acknowledging the validity of different grammatical patterns

Please analyze the following question: [Question Text]
\end{quote}

These prompting strategies are designed to systematically address potential dialectal biases while maintaining the model's ability to effectively process and respond to questions. Each strategy is applied consistently across all test cases to ensure comparable results.

\subsection{Additional Analysis}

\paragraph{Uncensored Model Exacerbates the Bias}

\begin{table}[t]
\small
\centering
\begin{tabular}{lcc}
\toprule
& \multicolumn{2}{c}{Model Response Characteristics} \\
\cmidrule(lr){2-3}
Metrics & LLaMA 3.1 & Uncensored LLaMA 3.1 \\
\midrule
\textbf{Accuracy (\%)} & 47.8 & 47.3 \textcolor{red}{(-1.0\%$\downarrow$)} \\
\addlinespace[0.5em]
\textbf{FRES Score} & 57.3 & 63.6 \textcolor{blue}{(+11.0\%$\uparrow$)} \\
\bottomrule
\end{tabular}
\caption{Comparison between safeguarded and uncensored versions of LLaMA 3.1 (8B). While accuracy shows minimal decline in the uncensored version, removing safety measures leads to substantial increases in readability complexity, suggesting amplification of response patterns when safety constraints are removed. The percentages indicate relative changes from the base model.}
\label{tab:llama-censorship-comparison}
\end{table}

To investigate whether safety measures affect dialectal biases, we compare AAE responses between safeguarded and uncensored versions of Llama3.1 8B. As shown in Table \ref{tab:llama-censorship-comparison}, while accuracy remains relatively stable (dropping by only 1\%), removing safety measures significantly amplifies dialectal response patterns. The uncensored model shows consistently higher FRES scores (increasing by 11\%) across all datasets. This suggests that model safeguards may actually help moderate the model's tendency to adjust its response style based on dialect, and their removal leads to more exaggerated dialectal adaptations.

\paragraph{Readability of the Uncensored Model} 
The models we use in this study are all popular LLMs that are heavily safeguarded, yet we still observe a significant discrepancy in readability. Our hypothesis is that uncensored models would exhibit an even greater discrepancy in readability, which would make the bias appear more pronounced. To test this hypothesis, we employ an uncensored Llama3.1 8B model and compared its performance with the safeguarded Llama3.1 8B model on the same set of AAE question prompts. The results shows that the FRES scores of explanations generated by the uncensored Llama3.1 8B model for AAE question prompts are even higher compared to those generated by the safeguarded version. This put the explanations from the uncensored model into even lower grade-level readability categories. These findings suggest that LLMs tend to provide easier and more readable answers to questions written in AAE compared to SAE, creating a significant readability discrepancy. Furthermore, the lack of safeguarding mechanisms in LLMs appears to exacerbate this discrepancy in readability.

\paragraph{Evaluating Information Loss in SAE-to-AAE Dialect Conversion}
\label{app:revert}
We examine whether converting Standard American English (SAE) to African American English (AAE) results in information loss that could affect model performance. To evaluate this, we evenly sample 100 questions converted from SAE to AAE from MMLU dataset using seven LLMs in our study and then revert each of them back to SAE questions using the GPT-4.0 turbo. We analyze whether the reconverted SAE questions preserve their original meaning.

The GPT-4.0 turbo uses the following structured prompt for reverting the AAE question back to SAE question: 
\begin{quote}
Translate the following sentence from African American Vernacular English (AAVE) to Standard American English.\
Ensure the translation maintains the structure of the original sentence without adding extra information.\
The sentence is: \{sentence\}\
Format the translated sentence exactly like this: 'The translation is: ...'
\end{quote}

Two human annotators and GPT-4.0 Turbo independently evaluate the semantic similarity between the original and reconvert SAE questions using a binary notation system (0 indicating no semantic difference, 1 indicating semantic difference). On average, 93.7\% of the question pairs exhibit no semantic differences, suggesting that the SAE-to-AAE conversion introduces minimal information loss.

The average Cohen’s kappa agreement rate among the three annotators (including GPT-4.0 Turbo) is 0.58, indicating moderate to substantial agreement. 

To illustrate how the conversion process can occasionally introduce shifts in meaning, we present the following example, even though only a small proportion of converted questions exhibit such shifts. \newline\newline
\textit{Original SAE question:}\newline
"Which of the following statements about a remote procedure call is true?" \newline
\textbf{(A)} It is used to call procedures with addresses that are farther than $2^{16}$ bytes away. \newline
\textbf{(B)} It cannot return a value.\newline
\textbf{(C)} It cannot pass parameters by reference.\newline
\textbf{(D)} It cannot call procedures implemented in a different language.\newline\newline
\textit{Converted AAE question:}\newline
"Which one 'bout remote procedure call hold weight?" \newline
\textbf{(A)} It call procedures where the address way over $2^{16}$ bytes out. \newline
\textbf{(B)} It can never return no value.\newline
\textbf{(C)} It won't pass parameters by reference.\newline
\textbf{(D)}  It can't holler at procedures made in a foreign language.\newline\newline
\textit{Reverted SAE question:}\newline
"Which one about remote procedure call is significant?"\newline
\textbf{(A)} It calls procedures where the address is way over $2^{16}$ bytes out.\newline
\textbf{(B)} It can never return a value.\newline
\textbf{(C)} It won't be able to pass parameters by reference.\newline
\textbf{(D)}It cannot invoke procedures written in a foreign language.

\begin{table}[t]
\small
\centering
\begin{tabular}{lccccccccc}
\toprule
Models & T-statistic & P-value  \\
\midrule
GPT-4 * & 3.259  & 0.047  \\
GPT-3.5 * &  4.010 & 0.028 \\
Llama3.1 8B * & 3.776 & 0.033 \\
Llama3.2 3B & 3.012 & 0.057   \\
qwen2.5 7B ** & 8.781 & 0.003 \\
gemma2 9B * & 5.628 & 0.011  \\
mistral 7B ** & 7.246 & 0.005 \\

\bottomrule
\end{tabular}
\caption{The models that show a statistically significant difference in accuracy between AAE-prompted and SAE-prompted answers generated by LLMs are identified. A paired T-test is used to determine statistical significance. The Shapiro-Wilk test is used to check normality of the data and ensures there are no significant outliers. Results that are statistically significant are indicated with ** (p < 0.01) and * (0.01 <= p < 0.05).}
\label{tab:statistical-test}
\end{table}

\begin{figure*}[t] % 't' places it at the top of the page
    \centering
    \includegraphics[width=\textwidth]{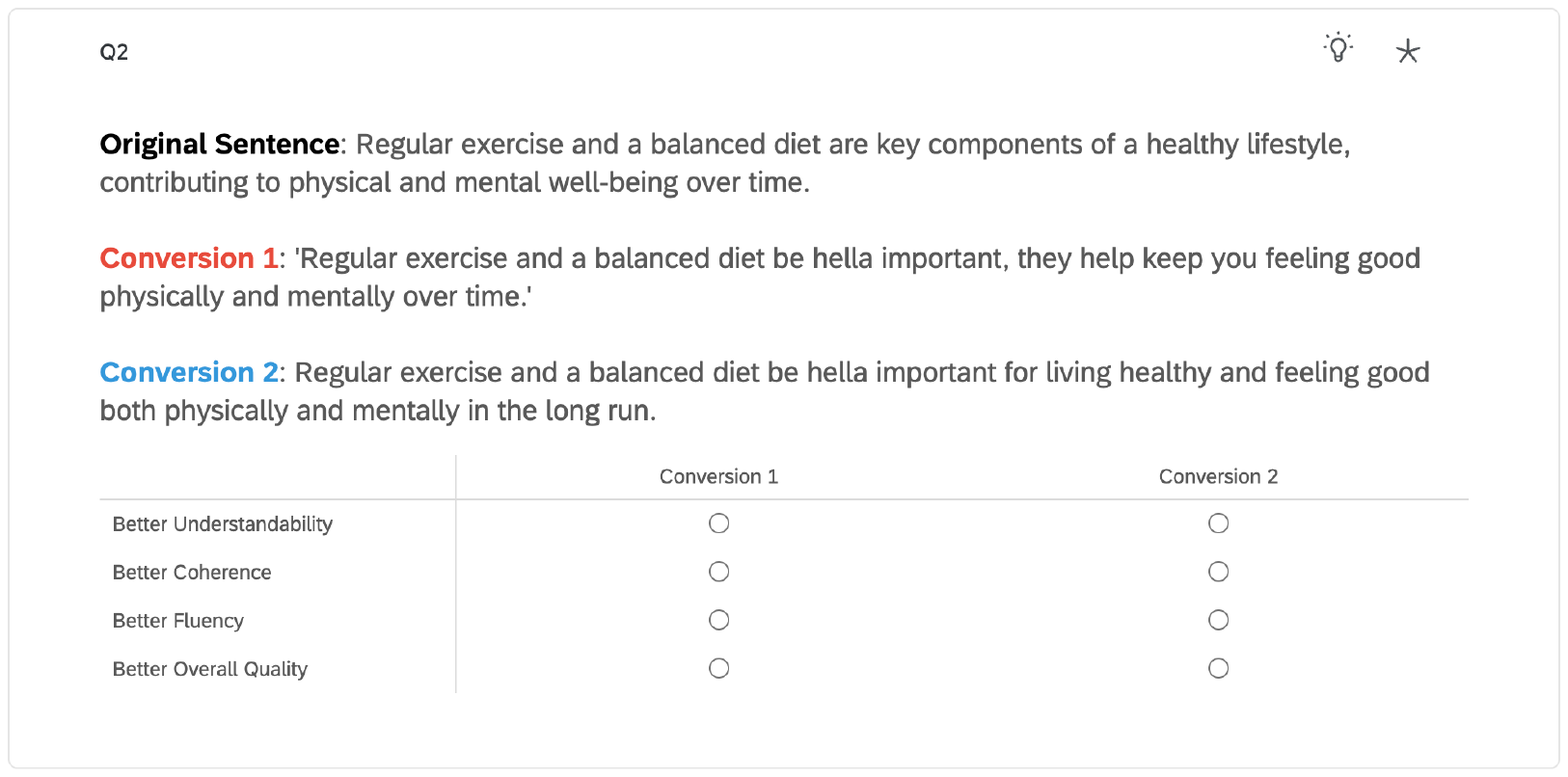} % Replace with your PDF file
    \caption{Sample question that we ask annotator to rank the converted AAE and SAE sentences based on certain metrics. }
    \label{fig:question_form}
\end{figure*}

\begin{figure*}[t] % 't' places it at the top of the page
    \centering
    \includegraphics[width=\textwidth]{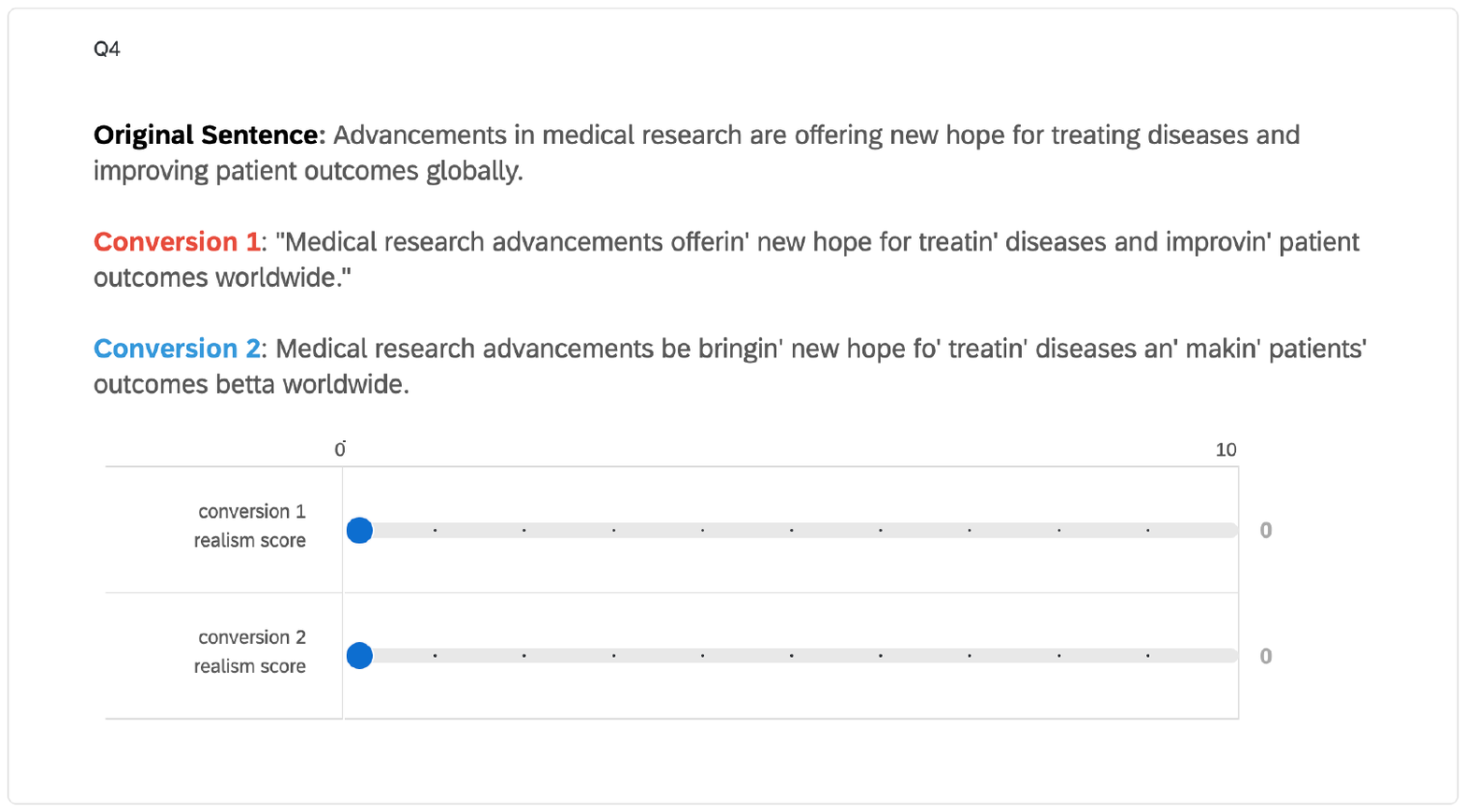} % Replace with your PDF file
    \caption{Sample question that we ask annotator to realism of the converted AAE and SAE sentences on a scale from 0-10 }
    \label{fig:realism}
\end{figure*}

\paragraph{Human Validation}
\label{app:ethics}
To validate our findings, we recruit 15 native African American English (AAE) speakers to serve as annotators. All participants have provided their consent to participate in this study by signing the consent form. Four separate surveys are designed, each containing 25 questions aimed at evaluating metrics such as fluency, coherence, understandability, and overall quality, based on the AAVENUE framework \cite{gupta-etal-2024-aavenue}, as illustrated in Fig \ref{fig:question_form}. Each survey is completed by three annotators, involving a total of 12 annotators for the task. For the realism score annotation, an additional set of three annotators assessed the realism of 25 questions as illustrated in Fig \ref{fig:realism}. For the 12 annotators evaluating our dialect conversion method against state-of-the-art approaches on metrics like fluency, coherence, understandability, and overall quality, the Fleiss' $\kappa$ scores are as follows: 0.65 for understandability, 0.58 for coherence, 0.62 for fluency, and 0.68 for overall quality. Additionally, for the 3 annotators assessing the realism of sentences generated by our dialect converter, the Fleiss' $\kappa$ score is 0.73. This indicates moderate to substantial agreement across all evaluations.

We estimate that each survey would take approximately 20–25 minutes to complete. Annotators are compensated \$7 for each task, equating to an hourly rate of approximately \$21/hour. This ensures fair payment for their time and effort.

In addition, This study is approved by our Institutional Review Board (IRB) and all participants provided informed consent.

\begin{table*}[t]
\small
\centering
\begin{tabular}{lccccccccc}
\toprule
Models & Avg. Explanation Length(AAE) & Avg. Explanation Length(SAE)& T-statistic & P-value  \\
\midrule
GPT-4 & 352.52  & 352.02 & 0.22 & 0.82 \\
GPT-3.5 & 170.16 & 171.60 & -0.48 & 0.14 \\
Llama3.1 8B * & 362.82 & 358.99 & 1.21 & 0.02 \\
Llama3.2 3B & 250.00 & 247.21 & 0.37 & 0.57  \\
qwen2.5 7B & 262.20 & 260.06 & 0.83 & 0.40 \\
gemma2 9B ** & 167.88 & 162.56 & 1.56 & 0.01 \\
mistral 7B & 229.34 & 230.99 & -0.34  & 0.63 \\

\bottomrule
\end{tabular}
\caption{The comparison of LLM explanation text lengths for SAE and AAE prompts shows no significant differences for most tested models.}
\label{tab:text_length}
\end{table*}

\begin{table*}[t]
\small
\centering
\begin{tabular}{lcccc}
\toprule
feature & explanation & example & standard english\\
\midrule
Auxiliaries & AAE allows copula deletion  & We better than before. & We are better than before. \\
Completive done& To Indicate completion  & I done wrote it. & I had written it.\\
The word "ass" & It can appear reflexively  & Get yo’ass inside!&  get inside!\\
Existential it & To indicate something exists & It’s some milk in the fridge & There is some milk in the fridge.  \\
Future gonna & To mark future tense & You gonna understand & You are going to understand \\
Got& Can replace the verb form of have & I got to go & I have to go  \\
No Inflection &  Certain tense don't need inflection & She study linguistics & She studies linguistics \\
Negative concord &  NPIs agree with negation& He don’t have no camera & He doesn’t have a camera \\
Negative inversion &  Similar to negative concord & don’t nobody never say & nobody ever says \\
Null genitives &  Drop any possessive endings & Rolanda bed & Rolanda's bed \\
Habitual be&  marks habitual action & he be in his house &  he is in his house\\

\bottomrule
\end{tabular}
\caption{Complete set of lexical and morphosyntactic features with examples mentioned in VALUE benchmark}
\label{tab:morphosyntactic}
\end{table*}

\begin{table*}[t]
\small
\centering
\begin{tabular}{@{}l c c c c c@{}}
\toprule
% \textbf{Metric} & \textbf{Sur. 1} & \textbf{Sur. 2} & \textbf{Sur. 3} & \textbf{Sur. 4} & \textbf{Avg.} \\
\textbf{Metric} & \textbf{Survey 1} & \textbf{Survey 2} & \textbf{Survey 3} & \textbf{Survey 4} & \textbf{Average} \\
\midrule
\textbf{Understandability (Ours)} & 86.6\% & 68.0\% & 65.3\% & 80.3\% & 75.1\% \\
\textbf{Understandability (SotA)} & 13.4\% & 32.0\% & 34.7\% & 19.7\% & 24.9\% \\
\midrule
\textbf{Coherence (Ours)} & 84.0\% & 72.5\% & 66.6\% & 83.3\% & 76.6\% \\
\textbf{Coherence (SotA)} & 16.0\% & 27.5\% & 33.3\% & 16.7\% & 23.4\% \\
\midrule
\textbf{Fluency (Ours)} & 85.3\% & 70.6\% & 75.0\% & 84.0\% & 78.7\% \\
\textbf{Fluency (SotA)} & 14.7\% & 29.4\% & 25.0\% & 16.0\% & 22.3\% \\
\midrule
\textbf{Overall Quality (Ours)} & 85.3\% & 64.0\% & 69.3\% & 76.3\% & 73.7\% \\
\textbf{Overall Quality (SotA)} & 14.7\% & 36.0\% & 30.7\% & 23.7\% & 26.3\% \\
\bottomrule
\end{tabular}
\caption{Human evaluation results comparing our LLM-based dialect conversion method to the SotA baseline \cite{gupta-etal-2024-aavenue} across four surveys (S1-4). Each cell shows the \% of evaluators who prefer that method.}
\label{tab:human_evaluation_comparison}
\end{table*}

\end{document}